\documentclass[11pt,a4paper]{article}
\pdfoutput=1
\usepackage[hyperref]{emnlp2020}
\usepackage{times}
\usepackage{latexsym}

\usepackage{microtype}
\usepackage{multirow}
\usepackage{hhline}
\usepackage{amsmath}
\newcommand{\bold}{\fontseries{b}\selectfont}

\aclfinalcopy 
\def\aclpaperid{1537} 

\newcommand{\comment}[1]{}

\title{Constrained Decoding for Computationally Efficient Named Entity Recognition Taggers}

\author{Brian Lester, Daniel Pressel, Amy Hemmeter, \\
{\bf Sagnik Ray Choudhury, and Srinivas Bangalore}
  \\ Interactions, Ann Arbor MI 48104 \\
  {\tt \{blester,dpressel,ahemmeter,schoudhury,sbangalore\}}\\
  {\tt@interactions.com} \\}

\date{}

\begin{document}
\maketitle
\begin{abstract}
Current state-of-the-art models for named entity recognition (NER) are neural models with a
conditional random field (CRF) as the final layer. Entities are represented as per-token labels
with a special structure in order to decode them into spans. Current work eschews prior knowledge 
of how the span encoding scheme works and relies on the CRF learning which transitions are
illegal and which are not to facilitate global coherence. We find that by constraining the
output to suppress illegal transitions we can train a tagger with a cross-entropy loss twice
as fast as a CRF with differences in F1 that are statistically insignificant, effectively
eliminating the need for a CRF. We analyze
the dynamics of tag co-occurrence to explain when these constraints are most effective and
provide open source implementations of our tagger in both PyTorch and TensorFlow.
\end{abstract}

\section{Introduction}

Named entity recognition (NER) is the task of finding phrases of interest in text that map to
real world entities such as organizations (``ORG'') or locations (``LOC''). This is normally cast as
a sequence labeling problem where each token is assigned a label that represents its entity type.
Multi-token entities are handled by having special ``Beginning'' and ``Inside'' indicators that
specify which tokens start, continue, or change the type of an entity. \citet{RatinovRothNER2009} show
that the IOBES tagging scheme, where entity spans must begin with a ``B'' token, end with an
``E`` token and where single token entities are labeled with an ``S'', performs better than the traditional
BIO scheme. The IOBES tagging scheme dictates that some token sequences are illegal. For example, one
cannot start an entity with an ``E'' tag (such as a transition from an ``O'', meaning it is outside of an entity,
to ``E-ORG'') nor can they change types in the middle of an entity---for example, transitioning from ``I-ORG'' to ``I-LOC''.
Most approaches to NER rely on the model learning which transitions are legal from the training data
rather than injecting prior knowledge of how the encoding scheme works.

It is conventional wisdom that, for NER, models with a linear-chain conditional random field (CRF)
\cite{Lafferty:2001:CRF:645530.655813} layer perform better than those without, yielding
relative performance increases between $2$ and $3$ percent in F1 \cite{ma-hovy:2016:P16-1:16, Lample2016NeuralAF:16}.
A CRF with Viterbi decoding promotes, but does not guarantee, global coherence while simple greedy decoding does not \cite{journals/jmlr/CollobertWBKKK11:11}. Therefore, in a bidirectional LSTM (biLSTM) model with a CRF layer, illegal
transitions are rare compared to models that select the best scoring tag for each token.

Due to the high variance observed in the performance of NER models \cite{reimers-gurevych-2017-reporting} it is important
to have fast training times to allow for multiple runs of these models. However, as the CRF forward algorithm is $O(NT^2)$,
where $N$ is the length of the sentence and $T$ is the number of possible tags, it slows down the training
significantly. Moreover, substantial effort is required to build an optimized, correct implementation of
this layer. Alternately, training with a cross-entropy loss runs in $O(N)$ for sparse labels and popular deep
learning toolkits provide an easy to use, parallel version of this loss which brings the runtime down to $O(\log N)$.

We believe that, due to the strong contextualized local features with infinite context created by today's neural models, global features
used in the CRF do little more than enforce the rules of an encoding scheme. Instead of traditional CRF training,
we propose training with a cross-entropy loss and using Viterbi decoding \cite{1450960} with heuristically determined transition
probabilities that prohibit illegal transitions. We call this constrained decoding and find that it allows us to
train models in half the time while yielding F1 scores comparable to CRFs.

\section{Method}

Training a tagger with a CRF is normally done by minimizing the negative log likelihood of the sequence of gold tags
given the input, parameterized by the model, where the probability of the sequence is given by

$$P(y|x;\theta) = \frac{e^{\sum_i\sum_j w_j f_j(y_{i-1}, y_i, x, i))}}{\sum_{y' \in Y} e^{\sum_i \sum_j w_j f_j(y'_{i - 1}, y'_{i}, x, i)}}$$

By creating a feature function, $f_j$, that is span-encoding-scheme-aware, we can introduce constraints that penalize
any sequence that includes an illegal transition by returning a large negative value. Note the
summation over all possible tag sequences. While efficient dynamic programs exist to make this sum
tractable for linear-chain CRFs with Markov assumptions, this is still a costly normalization factor to
compute.

In neural models, these feature functions are represented as a transition matrix that represents the score of
moving from one tag $y$ at index $i$ to another at $i + 1$. We implement a mask that effectively eliminates
invalid IOBES transitions by setting those scores to large negative values. By applying this mask to the
transition matrix we can simulate feature functions that down-weigh illegal transitions.

Contrast the CRF loss with the token-level cross-entropy loss where $y$ is the correct labels and $\hat{y}$
is the model's predictions.

\begin{align*}
    L_{\text{cross-entropy}} &= -\sum_i y_i \log(\hat{y}_i)
\end{align*}

Here we can see that the loss for each element in the input $i$ can be computed independently due to the
lack of a global normalization factor. This lack of a global view is potentially harmful, as we lose the
ability to condition on the previous label decision to avoid making illegal transitions. We hypothesize that,
using our illegal transition heuristics, we can create feature functions that do not have to be trained, but
can be applied at test time and allow for contextual coherence while using a cross-entropy loss.

We can use the mask directly as the transition matrix to calculate the maximum probability
sequence while avoiding illegal transitions for models that were not trained with a CRF.
Using these transitions scores in conjunction with cross-entropy trained models, we can achieve comparable models
that train more quickly. We call this method constrained decoding.

Constrained decoding is relatively easy to implement, given a working CRF implementation, all one needs to do is apply the transition mask to the CRF transition parameters to create a constrained CRF. Replacing the transition parameters with the mask yields our constrained decoding model. Starting from scratch, one only needs to implement Viterbi decoding, using the mask as transition parameters, to implement the constrained decoding model---avoiding the need for the CRF forward algorithm and the CRF loss.

For constrained decoding, we leverage the IOBES tagging scheme rather than BIO tagging, allowing us to
inject more structure into the decoding mask. Early experiments with BIO tagging failed to show the
large gains we realized using IOBES tagging for the reasons mentioned in Section \ref{section:Analysis}.

\section{Experiments \& Results}

To test if we can replace the CRF with constrained decoding we use two sequential prediction tasks:
NER (CoNLL 2003 \cite{TjongKimSang:2003:ICS:1119176.1119195}, WNUT-17 \cite{Derczynski2017ResultsOT},
and OntoNotes \cite{Hovy:2006:O9S:1614049.1614064}) and slot-filling (Snips \cite{Coucke2018SnipsVP}).
For each (task, dataset) pair we use common embeddings and hyperparameters from the literature. The baseline
models are biLSTM-CRFs with character compositional features based on convolutional neural networks \cite{DosSantos:2014:LCR:3044805.3045095:14} and our models are identical except we train with a cross-entropy
loss and use the encoding scheme constraints as transition probabilities instead of learning them with
a CRF. Our hyper-parameters mostly follow \citet{ma-hovy:2016:P16-1:16}, except we use multiple pre-trained word embeddings concatenated together \cite{lester2020multiple}. For Ontonotes we follow \citet{Chiu2016NamedER}. See Section \ref{sec:hp} or the configuration files in our implementation for more details.

As seen in Table \ref{tab:constrain-results}, in three out of four datasets constrained decoding performs comparably
or better than the CRF in terms of F1. OntoNotes
is the only dataset with a statistically significant difference in performance. We explore this discrepancy
in Section \ref{section:Analysis}. Similarly, Table \ref{tab:internal-results} shows that when we apply
constrained decoding to a variety of internal datasets, which span a diverse set of specific domains, we do
not observe any statistically significant differences in F1 between CRF and constrained decoding models.

The models were trained using Mead-Baseline \cite{Baseline:2018}, an open-source framework for creating, training, evaluating and deploying models for NLP. The constrained decoding tagger performs much faster at training time. Even when compared to the optimized, batched CRF provided by Mead-Baseline, it trained in 51.2\% of the time as the CRF.

In addition to faster training times, training our constrained models produces only 65\% of the $\text{CO}_2$ emissions that the CRF does. While GPU computations for the constrained model draw $1.3$ times more power---due to the greater degree of possible parallelism in the cross-entropy loss function---than the CRF, the reduction in training time results in smaller carbon emissions as calculated in \citet{strubell-etal-2019-energy}.

\begin{table}[t]
\centering
\begin{tabular}{l l | r r r r}
    Dataset & Model & mean & std & max \\
    \hhline{==|====}
    CoNLL & CRF & \bold{91.61} & 0.25 & \bold{92.00} \\
    & Constrain & 91.44 & 0.23 & 91.90 \\
    \hline
    WNUT-17 & CRF & 40.33 & 1.13 & \bold{41.99} \\
    & Constrain & \bold{40.59} & 1.06 & 41.71 \\
    \hline
    Snips & CRF & 96.04 & 0.28 & \bold{96.35} \\
    & Constrain & \bold{96.07} & 0.17 & 96.29 \\
    \hline
    OntoNotes & CRF & \bold{87.43} & 0.26 & \bold{87.57} \\
    & Constrain & 86.13 & 0.17 & 86.72 \\
    
\end{tabular}
\caption{
    Tagging results on a variety of datasets. The CRF model is a
    standard biLSTM-CRF while the Constrain model is a biLSTM trained with a cross-entropy loss that
    uses heuristic transition scores, created from the illegal transitions, for test time decoding. OntoNotes
    is the only dataset where the difference in performance between the CRF and constrained decdoing is statistically
    significant $(p < 0.5)$.
    All scores are entity-level F1 and are reported across 10 runs.
}
\label{tab:constrain-results}
\end{table}

\begin{table}[t]
\centering
\begin{tabular}{l l | r}
    Task & Domain & $\Delta$ \\
    \hhline{==|=}
    NER & Generic NER & 0.80 \\
    Slot Filling & Customer Service & 0.21 \\
                 & Automotive & -0.68 \\
                 & Cyber Security & 0.84\\
\end{tabular}
\caption{
    Entity-level F1 comparing a constrained CRF model with a constrained decoding model.
    Due to the nature of the the data we present the relative performance difference between the
    two models. We see some improvements and some drops in performance but, once again, there is not a statistically
    significant difference between the CRF and constrained decoding.
}
\label{tab:internal-results}
\end{table}

\begin{table}[t]
\centering
\begin{tabular}{l l | r}
    Task & Dataset & $\Delta$ \\
    \hhline{==|=}
    NER & CoNLL & -0.03 \\
        & WNUT-17 & 0.65 \\
        & OntoNotes & -1.48 \\
        & Snips & 0.03 \\
\end{tabular}
\caption{
    Results on well-known datatsets presented as relative differences to help frame results in Table \ref{tab:internal-results}
}
\label{tab:relative-results}
\end{table}

Constrained decoding can also be applied to a CRF. The CRF does not always learn the rules
of a transition scheme, especially in early training iterations. Applying the constraints to the CRF can improve
both F1 and convergence speed. We establish this by training biLSTM-CRF models with and without constraints
on CoNLL 2003. We find that the constraint mask yields a small (albeit statistically insignificant) boost in F1 as shown in Table \ref{tab:constrain-crf}.

\begin{table}
\centering
\begin{tabular}{l | r r r}
    Model & mean & std & max \\
    \hhline{=|===}
    Unconstrained & 91.55 & 0.26 & 91.79 \\
    Constrained & \bold{91.61} & 0.25 & \bold{92.00} \\
\end{tabular}
\caption{
    Results of biLSTM-CRF models with and without constraints evaluated with entity-level F1 on the
    CoNLL 2003 dataset. Scores are reported across 10 runs. We see that while, in theory, the CRF
    should learn the constraints, injecting this knowledge gives a gain in performance.
}
\label{tab:constrain-crf}
\end{table}

Our experiments suggest that injecting prior knowledge of the transition scheme helps the model to focus on learning the features for sequence tagging tasks (and not the transition rules themselves) and train faster. Table \ref{tab:converge} shows that our constrained model converged
\footnote{We define convergence as the epoch where development set performance stops improving} on CoNLL 2003
faster on average than an unconstrained CRF.

\begin{table}
\centering
\begin{tabular}{l | r r r r}
    Model & mean & std & min & max \\
    \hhline{=|====}
    Unconstrained & 72.4 & 21.0 & 16 & 97 \\
    Constrained & \bold{60.6} & 23.3 & 37 & 89 \\
\end{tabular}
\caption{
    Using the constraints while training a biLSTM-CRF tagger on the CoNLL dataset result in a statistically
    significant $(p < 0.5)$ decrease in the number of epochs until convergence. Scores are reported across 30 runs.
}
\label{tab:converge}
\end{table}

\section{Analysis}
\label{section:Analysis}

\begin{table*}[ht]
\centering
\begin{tabular}{l | r r r r r}
    Dataset & Tag Types & Ambiguity & Strictly Dominated & Easy First & Easy Last \\
    \hhline{=|=====}
    CoNLL (IOBES) &  4 &  8.8\% & 71.2\% & 58.3\% & 94.0\% \\
    CoNLL (BIO)   &  4 &  7.4\% & 59.6\% & 68.5\% & 57.4\% \\
    WNUT-17       &  6 &  3.6\% & 74.3\% & 82.9\% & 97.0\% \\
    OntoNotes     & 18 & 14.9\% & 15.9\% & 16.2\% & 55.9\% \\
    Snips         & 39 & 24.5\% & 26.7\% & 32.4\% & 91.1\% \\
\end{tabular}
\caption{
    Analysis of the tag dynamics and co-occurrence. We see that OntoNotes is an outlier in the percentage of
    ambiguous tokens that are strictly dominated by their context, the entities that have easy to spot starting tokens,
    and entities with clearly defined ends. All of these quirks of the data help explain why we only see a statistically
    significant performance drop for OntoNotes.
}
\label{tab:tag-cooccur}
\end{table*}


The relatively poor performance of constrained decoding on OntoNotes suggests that there are several classes of
transition that it cannot model. For example, the  transition distribution between entity types, or the prior
distribution of entities. We analyzed the datasets to identify the characteristics that cause constrained decoding to fail.

One such presumably obvious characteristic is the number of entity types. However, our experiments suggest that number of entity types does not affect performance: Snips has more entity types than OntoNotes yet constrained decoding works better for Snips. 

We define an ambiguous token as a token whose type has multiple tag values in the dataset. For example the token ``Chicago'' could be ``I-LOC'' or ``I-ORG'' in the phrases ``the Chicago River'' and ``the Chicago Bears'' respectively. Such ambiguous tokens are the ones for which we expect global features to be particularly useful. A ``strictly dominated token'' is defined as a token that can only take on a single value due to the legality of the transition from the previous tag. In the above example given that ``the'' was a ``B-LOC'' then
``Chicago'' is strictly dominated and forced to be an ``I-LOC'.
Contrast this with a non-strictly dominated token that can still have multiple possible tag values when conditioned on
the previous tag.
As constrained decoding eliminates illegal transitions
we would expect that it would perform well on datasets where a large proportion of ambiguous tokens are strictly dominated. This tends to hold true---only $15.9$\% of OntoNotes' ambiguous tokens are strictly dominated while $70.7$\% of CoNLL's tokens are and for WNUT-17 $73.6$\% are.

We believe that the ambiguity of the first and last token of an entity also plays a role. Once we start an entity, constrained decoding vastly narrows the scope of decisions that need to be made. Instead of making a decision over the entire set of tags, we only decide if we should continue the entity with an ``I-'' or end it with an ``E-''. Therefore, we expect constrained decoding to work well with datasets that have fairly unambiguous entity starts and ends. We quantify this by finding the proportion of entities that begin (or end) with an unambiguous type, that is, the first token of an entity only has a single label throughout the dataset, for example, ``Kuwait'' is only labeled with ``S-LOC'' in the CoNLL dataset. We call these metrics ``Easy First'' and ``Easy Last'' respectively and find that datasets with higher constrained decoding performance also have a higher percentages of entities with an easy first or last token. A summary of these characteristics for each dataset is
found in Table \ref{tab:tag-cooccur}.

This also explains why constrained decoding doesn't work as well for BIO-encoded CoNLL as it does for IOBES. When using the IOBES format, more tokens are strictly dominated. The other stark difference is the proportion of ``Easy Last'' entities. Without the ``E-'' token, much less structure can be injected into the model, resulting in decreased performance of constrained decoding. These trends also hold true in internal datasets, where the Automotive dataset had the fewest
incidences of each of these phenomena.

While not perfect predictors for the performance of constrained decoding, the metrics chosen are good proxies and can be used as a prescriptive measure for new datasets.

\section{Previous Work}

Our approach is similar in spirit to previous work in NLP where constraints are introduced during
 training and inference time \cite{Roth:2005:ILP:1102351.1102444, Punyakanok:2005:LIO:1642293.1642473} to lighten the
computational load, and to \citet{strubell-etal-2018-linguistically} where prior knowledge
is injected into the model by manual manipulation. In our approach, however, we focus specifically on manipulating the model weights themselves rather than model features.

There have been attempts to eliminate the CRF layer, notably, \citet{shen2018deep} found
that an additional LSTM greedy decoder layer is competitive with the CRF layer, though their baseline is much weaker than the models found in other work. Additionally, their decoder has an auto-regressive relationship that is difficult to parallelize and, in practice, there is still significant overhead at training time. \citet{Chiu2016NamedER} mention good results with a similar technique but don't provide in-depth analysis, metrics, or test its generality.

\section{Conclusion}

For sequence tagging tasks, a CRF layer introduces substantial computational cost. We propose replacing it with a lightweight technique, constrained decoding, which doubles the speed of training with comparable F1 performance. We analyze the algorithm to understand where it might work or fail and propose prescriptive measures for using it.

The broad theme of the work is to find simple and computationally efficient modifications of current networks and suggest
possible failure cases. While larger models have shown significant improvements, we believe there is still relevance in investigating small, targeted changes. In the future, we want to explore similar techniques in other common NLP tasks.

\bibliography{emnlp2020}
\bibliographystyle{acl_natbib}

\appendix

\section{Reproducibility}

\subsection{Hyperparameters}

Mead/Baseline is a configuration file driven model training framework. All hyperparameters are fully specified in
the configuration files included with the source code for our experiments.

\subsection{Statistical Significance}

For all claims of statistical significance we use a t-test as implemented in \texttt{scipy} \cite{2020SciPy-NMeth} and using an alpha value of $0.05$.

\subsection{Computational Resources}

All models were trained on a single NVIDIA 1080Ti. While multiple GPUs were used for training many models in parallel
to facilitate testing many datasets and to estimate the variability of the method, the actual model can easily be trained
on a single GPU.

\subsection{Evaluation}

To calculate metrics, entity-level F1 is used for NER and slot-filling. In entity-level F1, entities are created from the token-level labels and compared to the gold entities. Entities that
match on both type and boundaries are considered correct while a mismatch in either causes an
error. The F1 score is then calculated using these entities.
We use the evaluation code that ships with the framework we use, MEAD/Baseline, which we have bundled
with the source code for our experiments.

\subsection{Model Size}

The number of parameters in different models can be found in Table \ref{tab:parameter-stats}.

\begin{table}[t]
\centering
\begin{tabular}{l l | r }
    Dataset & Model & Parameters \\
    \hhline{==|=}
    CoNLL & CRF & 4,658,190 \\
    & Constrain & 4,657,790 \\
    & Unconstrained CRF & 4,658,190 \\
    \hline
    WNUT-17 & CRF & 12,090,032 \\
    & Constrain & 12,089,248 \\
    \hline
    Snips & CRF & 5,940,866 \\
    & Constrain & 5,924,737 \\
    \hline
    OntoNotes & CRF & 12,090,032 \\
    & Constrain & 12,089,248 \\
    
\end{tabular}
\caption{The number of parameters for different models.}
\label{tab:parameter-stats}
\end{table}

\subsection{Dataset Information}

Relevant information about datasets can be found in Table \ref{tab:data-stats}. The majority of data is used as distributed,
except we convert NER and slot-filling datasets to the IOBES format. All public datasets are included in the supplementary
material. A quick overview of each dataset follows:

\textbf{CoNLL}: A NER dataset based on news text. We converted the IOB labels into the IOBES format. There are 4 entity types, \texttt{MISC}, \texttt{LOC}, \texttt{PER}, and \texttt{LOC}.

\textbf{WNUT-17}: A NER dataset of new and emerging entities based on noisy user text. We converted the BIO labels into the IOBES format. There are 6 entity types, \texttt{corporation},
\texttt{creative-work}, \texttt{group}, \texttt{location}, \texttt{person}, and \texttt{product}.

\textbf{OntoNotes}: A much larger NER dataset. We converted the labels into the IOBES format. There are 18 entity types, \texttt{CARDINAL}, \texttt{DATE}, \texttt{EVENT}, \texttt{FAC}, \texttt{GPE}, \texttt{LANGUAGE}, \texttt{LAW}, \texttt{LOC}, \texttt{MONEY}, \texttt{NORP}, \texttt{ORDINAL}, \texttt{ORG}, \texttt{PERCENT}, \texttt{PERSON}, \texttt{PRODUCT}, \texttt{QUANTITY}, \texttt{TIME}, and \texttt{WORK\_OF\_ART}.

\textbf{Snips}: A slot-filling dataset focusing on commands one would give a virtual assistant. We converted the dataset from its normal format of two associated files, one containing surface terms and one containing labels in the more standard CoNLL file format and converted the labels into the IOBES format. There are 39 entity types, \texttt{album}, \texttt{artist}, \texttt{best\_rating}, \texttt{city}, \texttt{condition\_description}, \texttt{condition\_temperature}, \texttt{country}, \texttt{cuisine}, \texttt{current\_location}, \texttt{entity\_name}, \texttt{facility}, \texttt{genre}, \texttt{geographic\_poi}, \texttt{location\_name}, \texttt{movie\_name}, \texttt{movie\_type}, \texttt{music\_item}, \texttt{object\_location\_type}, \texttt{object\_name}, \texttt{object\_part\_of\_series\_type}, \texttt{object\_select}, \texttt{object\_type}, \texttt{party\_size\_description}, \texttt{party\_size\_number}, \texttt{playlist}, \texttt{playlist\_owner}, \texttt{poi}, \texttt{rating\_unit}, \texttt{rating\_value}, \texttt{restaurant\_name}, \texttt{restaurant\_type}, \texttt{served\_dish}, \texttt{service}, \texttt{sort}, \texttt{spatial\_relation}, \texttt{state}, \texttt{timeRange}, \texttt{track}, and \texttt{year}.

\begin{table*}[t]
    \centering
    \begin{tabular}{l l | r r r r}
        Dataset & & Train & Dev & Test & Total \\
        \hhline{==|====}
        CoNLL & Examples & 14,987 & 3,466 & 3674 & 22137 \\
              & Tokens & 204,567 & 51,578 & 46,666 & 302,811 \\
        WNUT-17 & Examples & 3,394 & 1,009 & 1,287 & 5,690 \\
                & Tokens & 62,730 & 15,733 & 23,394 & 101,857 \\
        OntoNotes & Examples & 59,924 & 8,528 & 8,262 & 76,714 \\
                  & Tokens & 1,088,503 & 147,724 & 152,728 & 1,388,955 \\
        Snips & Examples & 13,084 & 700 & 700 & 14,484  \\
              & Tokens & 117,700 & 6,384 & 6,354 & 130,438 \\
    \end{tabular}
    \caption{Example and token count statistics for public datasets used.}
    \label{tab:data-stats}
\end{table*}

\subsection{Hyper Parameters}
\label{sec:hp}

Table \ref{tab:hps} details the various hyper-parameters used to train models for each dataset. For all datasets the only difference between the baseline CRF model and the model using constrained decoding is that the CRF has learnable transition parameters in the final layer while the constrained decoding model sets these transitions parameters manually based on the rules of the span encoding scheme. The framework we use, Mead-Baseline, is configuration file driven and we have included the configuration files used on our experiments in the supplementary material.

\begin{table*}[t]
    \centering
    \begin{tabular}{l|r r r r}
        HyperParameter & CoNLL & Ontonotes & Snips & WNUT-17 \\
        \hhline{=|====}
        Embedding & 6B + Senna & 6B + Senna & 6B + GN & 27B + w2v-30M + 840B \\
        Character Filter Size & 3 & 3 & 3 & 3 \\
        Character Feature Size & 30 & 30 & 30 & 30 \\
        Character Embed Size & 30 & 20 & 30 & 30 \\
        RNN Type & biLSTM & biLSTM & biLSTM & biLSTM \\
        RNN Size & 400 & 400 & 400 & 200 \\
        RNN Layers & 1 & 2 & 1 & 1 \\
        Drop In & 0.1 & 0.1 & 0.1 & 0.0 \\
        Drop Out & 0.5 & 0.63 & 0.5 & 0.5 \\
        Batch Size & 10 & 9 & 10 & 20 \\
        Epochs & 100 & 100 & 100 & 60 \\
        Learning Rate & 0.015 & 0.008 & 0.015 & 0.008 \\
        Momentum & 0.9 & 0.9 & 0.9 & 0.9 \\
        Gradient Clipping & 5.0 & 5.0 & 5.0 & 5.0 \\
        Optimizer & SGD & SGD & SGD & SGD \\
        Patience & 40 & 40 & 40 & 20 \\
        Early Stopping Metric & f1 & f1 & f1 & f1 \\
        Span Type & IOBES & IOBES & IOBES & IOBES 
    \end{tabular}
    \caption{
        Hyper-parameters used for each dataset. ``Embedding'' is the type of pre-trained word embeddings used. 6B, 27B, and 840B are GloVe embeddings \cite{pennington2014glove} with 27B having been trained on Twitter, Senna is embeddings from \citet{journals/jmlr/CollobertWBKKK11:11}, GN is vectors trained on Google News with word2vec from \citet{mikolovgn} and w2v-30M are word2vec vectors trained on Twitter from \citet{Baseline:2018}. ``Character Filter Size'' is the number of token the character compositional convolutional neural network cover is a single window, ``Character Feature Size'' is the number of convolutional features maps used, and ``Character Embed Size'' is the dimensionality of the vectors each character is mapped to before it is the input to the convolutional network. The ``RNN Size'' is the size of the output after the RNN which means that bidirectional RNNs are composed to two RNNs, one in each direction, where both are half the ``RNN Size''. ``Drop In'' is the probability that an entire token will be drop out from the input, while ``Drop Out'' is the probability that individual neurons are dropped out \cite{JMLR:v15:srivastava14a}.
    }
    \label{tab:hps}
\end{table*}
\end{document}